\documentclass[10pt,twocolumn,letterpaper]{article}

\usepackage{cvpr}              

\usepackage{graphicx}
\usepackage{amsmath}
\usepackage{amssymb}
\usepackage{booktabs}
\usepackage{tabularx}
\usepackage[accsupp]{axessibility}

\usepackage[pagebackref,breaklinks,colorlinks]{hyperref}

\usepackage[capitalize]{cleveref}
\crefname{section}{Sec.}{Secs.}
\Crefname{section}{Section}{Sections}
\Crefname{table}{Table}{Tables}
\crefname{table}{Tab.}{Tabs.}

\def\cvprPaperID{2} 
\def\confName{CVPR}
\def\confYear{2022}

\begin{document}

\title{Unsupervised domain adaptation and super resolution on drone images for autonomous dry herbage biomass estimation}
\author{Paul Albert$^{1,3,5}$, Mohamed Saadeldin$^{2,3,5}$, Badri Narayanan$^{2,3,5}$, Jaime Fernandez$^{1,3}$, Brian Mac Namee$^{2,3,5}$, \\ Deirdre Hennessy$^{4,5}$, Noel E. O{'}Connor$^{1,3,5}$, Kevin McGuinness$^{1,3,5}$\\\\
$^1$School of Electronic Engineering, Dublin City University\\ $^2$School of Computer Science, University College Dublin\\ $^3$Insight Centre for Data Analytics \\ $^4$Teagasc, $^5$VistaMilk\\
{\tt\small paul.albert@insight-centre.org}
}
\maketitle

\begin{abstract}
   Herbage mass yield and composition estimation is an important tool for dairy farmers to ensure an adequate supply of high quality herbage for grazing and subsequently milk production. By accurately estimating herbage mass and composition, targeted nitrogen fertiliser application strategies can be deployed to improve localised regions in a herbage field, effectively reducing the negative impacts of over-fertilization on biodiversity and the environment. In this context, deep learning algorithms offer a tempting alternative to the usual means of sward composition estimation, which involves the destructive process of cutting a sample from the herbage field and sorting by hand all plant species in the herbage. The process is labour intensive and time consuming and so not utilised by farmers. Deep learning has been successfully applied in this context on images collected by high-resolution cameras on the ground. Moving the deep learning solution to drone imaging, however, has the potential to further improve the herbage mass yield and composition estimation task by extending the ground-level estimation to the large surfaces occupied by fields/paddocks. Drone images come at the cost of lower resolution views of the fields taken from a high altitude and requires further herbage ground-truth collection from the large surfaces covered by drone images. This paper proposes to transfer knowledge learned on ground-level images to raw drone images in an unsupervised manner. To do so, we use unpaired image style translation to enhance the resolution of drone images by a factor of eight and modify them to appear closer to their ground-level counterparts. We then use the enhanced drone images to train a semi-supervised algorithm that uses ground-truthed, ground-level images as the labelled data together with a large amount of unlabeled drone images. We validate our results on a small held-out drone image test set to show the validity of our approach, which opens the way for automated dry herbage biomass monitoring~\url{www.github.com/PaulAlbert31/Clover_SSL}.
\end{abstract}

\begin{figure}[t]
\centering{}\includegraphics[width=\linewidth]{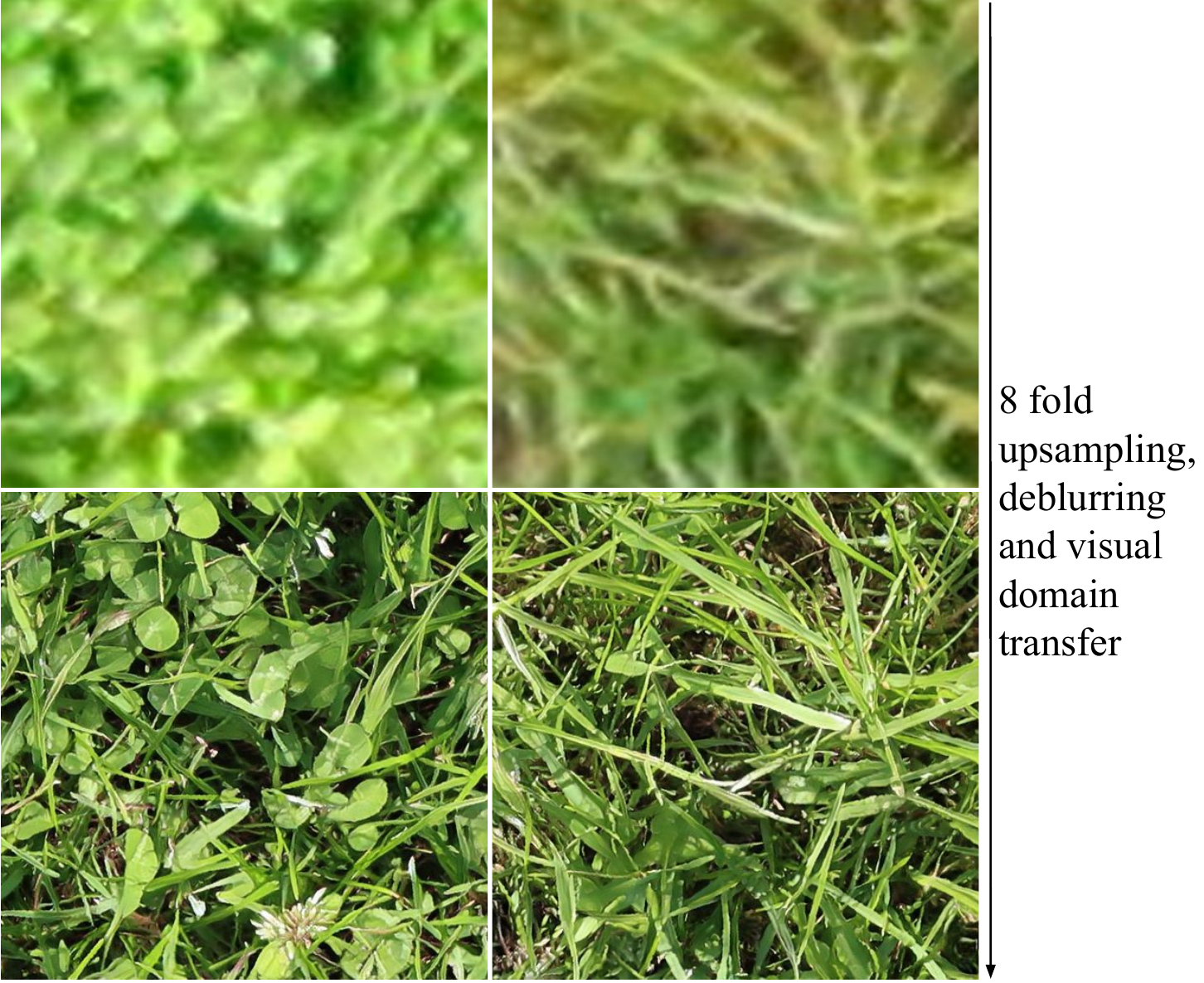}
\caption{Up-sampling drone images by a factor of 8. Images at the top are $64\times64$ crops from drone images. Images at the bottom are up-sampled to $512\times512$, deblurred and transferred to the ground-level visual domain in an unpaired fashion. We use the transformed images in a semi-supervised regression objective.~\label{fig:upsamplingexample}}
\end{figure}

\section{Introduction}
\label{sec:intro}
Nitrogen fertilization has proven to be efficient in enhancing herbage quantity and quality, yet over-fertilization has detrimental effects on biodiversity and on the environment in general~\cite{2016_SH_overfertreview,2004_AMBIO_nitrogenchina,2015_JSSPN_fertcentralchile}. In this context, clover proves to be an important ally to the farmer for two reasons. First, clover naturally captures widely available nitrogen from the atmosphere and renders it available in the soil for the grass to use~\cite{2009_AAS_nitrogen, 2009_JAE_biomassMixture}. Second, having proper amounts of clover in the feed has been shown to increase cow appetite, which in turn translates to higher milk production~\cite{2017_JOI_biomassMixture2,2018_JDS_incorporatingVM}. Monitoring clover content in the herbage then becomes an important aspect of milk production and regular herbage biomass probing is performed by humans to ensure a proper grass to clover balance. The herbage probing process involves cutting a sample from the field, drying it in lab before manually separating each component of the herbage by hand~\cite{2018_JDS_incorporatingVM}. Knowing the herbage composition and dry mass is valuable for the farmer but because existing probing processes are destructive and time consuming it is never probed at the farm level. In this context, deep learning has the capacity to provide a simpler, non-destructive alternative to dry herbage phenotyping and mass estimation from images alone. The feasibility of the method has been shown in previous works, often relying on partially labeled or unlabeled images to reduce the strain of the lengthy data collection process~\cite{2020_IMVIP_extracting,2021_ICCVW_semisupmine,2019_CVPRW_grasscloverdataset}. These works, however, only studied the application of deep learning algorithms to ground-level images using handheld devices and tripods~\cite{2021_EGF_Irishdataset} or all terrain vehicles (ATV)~\cite{2018_ICPA_grasscloverfirst}. 
In this paper, we propose to extend the dry biomass and herbage mass estimation problem to drone images, which are more suitable for covering large herbage fields. Because drones operate at higher altitudes, large land areas that can span from tens to hundreds of square meters depending on the altitude are captured in every drone image, rendering the fine ground-truthing of data very challenging. To mitigate this issue, we propose to transfer knowledge learned from few high-resolution ground-level images to drone images in an unsupervised manner. To do so, we apply an unpaired domain transfer algorithm~\cite{2020_ECCV_CUT} to the drone images to enhance their resolution to $2048\times 2048$ and to reduce the visual domain gap with the ground-level images (see Figure~\ref{fig:upsamplingexample}). We then train a semi-supervised neural network for regression on a small number of labeled ground-level images together with unlabeled drone images to effectively transfer knowledge between the two domains. To evaluate the quality of our regression algorithm, we test it on a small data set of ground-truthed drone images collected in Ireland and evaluate the benefit the large quantities of unlabeled images drone imagery provides to improve the ground-level predictions. Our contributions are:
\begin{enumerate}
    \item 328 drone images of herbage fields in Ireland;
    \item An unpaired image transfer pipeline, increasing the resolution of drone images 8 fold and transferring them to the ground-level camera visual domain;
    \item A semi-supervised regression that learns to estimate dry herbage biomass from a small set of annotated ground-level images and unlabeled drone images.
\end{enumerate}

\section{Related work}
\subsection{Computer vision for agriculture}
Computer vision offers possibilities to revolutionize smart agriculture by providing farmers with automated solutions to address deficiencies in their fields and also to drive efficiencies in their daily practice. Weed detection, for example, is a topic that received significant attention where undesirable weeds are automatically detected in a field using image analysis. The weed detection process encompasses simple edge detection or color filtering approaches~\cite{2016_RFIUA_weeddetectionsoil,2016_IRJET_weeddetection,2016_CEA_weeddetectchina}; random forest classifiers trained on color features~\cite{2020_KSKD_weeddetectchiliRF}; or more recently semantic segmentation neural networks~\cite{2017_ICICTI_weedsegcnn}. Other popular phenotyping tasks include fruit detection and counting in trees~\cite{2020_FPS_tomato,2016_Sensors_deepfruits,2019_FPS_singleshotpear} or wheat head identification~\cite{2020_PP_globalwheatdataset}.
Generative adversarial networks (GANs) are of special interest in the plant domain. We separate here GANs architectures between the conditional architectures that are trained with pairs of input and outputs in the two different visual domains~\cite{2017_CVPR_conditionalGAN} and unpaired architectures, i.e. CycleGAN~\cite{2017_ICCV_cyclegan} or Contrastive Unpaired Translation~\cite{2020_ECCV_CUT} where images from both visual domains are not semantically linked. Conditional GANs have been successfully applied to generate RGB images from semantic segmentation masks~\cite{2018_BMVC_cGanleaf}, to predict cabbage growth~\cite{2021_CEA_GANcabage}, or plant super-resolution to improve feature detection~\cite{2020_IEEE_leafsuperres}. Unpaired GANs have been used to estimate disease spreading on leaves~\cite{2018_CCC_GANdisease2,2020_CEA_Gandisease,2019_SMJ_GANdiseasegen} or to improve the realism of synthetic images~\cite{2020_CEA_cycleganfruits,2021_Plants_CycleGANwheat}. Finally, drone (UAV) imaging holds important potential for automating farm tasks since drones can easily cover large areas of uneven terrain~\cite{2020_PB_phenotypingreview}. Deep learning has been successfully applied to derive growth rate from nitrogen fertilization on drone images~\cite{2016_RS_DroneWheat}, estimate the emergence rate of seeds in the field~\cite{2017_RS_seedemergeanceUAV}, wheat density~\cite{2017_RSE_plantdensityUAV}, weed detection~\cite{2019_AAGSSAOP_weedsuav}, and land classification~\cite{2020_CVPR_AGVISchallenge}. The main drawback when applying deep learning on drone images for plant phenotyping remains the difficulty of ground-truthing the images because of the large areas covered~\cite{2020_PB_phenotypingreview}.

\subsection{Biomass composition prediction from canopy images}
Herbage biomass composition from images gained traction after the publication of the GrassClover image dataset for semantic and hierarchical species understanding in agriculture~\cite{2018_ICPA_grasscloverfirst,2019_CVPRW_grasscloverdataset}. To solve the biomass composition problem, Skovsen~\etal~\cite{2019_CVPRW_grasscloverdataset} propose to create artificial images where grass/clover/weeds elements are manually cropped from the raw images and pasted in a random fashion on a soil background image to create a synthetic but fully segmented image. A semantic segmentation network is then trained on the synthetic data and predicts species pixel percentages from the real RGB images and a least-squares regression algorithm predicts the dry biomass from the pixel percentages. The Irish grass clover dataset~\cite{2021_EGF_Irishdataset} proposes, additionally to the dry biomass percentages, to predict the herbage height pre-grasing (cm) and the dry matter per hectare (kg DM/ha) from the canopy images. Although both datasets provide an additional large amount of unlabeled images, the respectively baseline are purely supervised and do not make use of the raw images. Subsequent algorithms were published and tested on both datasets to attempt to use the raw data to improve the biomass prediction. Narayannan~\etal~\cite{2020_IMVIP_extracting} proposed to use mean imputation to infer labels for the partially labeled samples before training a convolutional neural network (CNN) on the larger dataset. Albert~\etal~\cite{2021_ICCVW_semisupmine} generate synthetic semantic segmentation images in a similar fashion to Skovsen~\etal~\cite{2019_CVPRW_grasscloverdataset} but instead of using the linear regression algorithm to predict at test time, the regressor is used to automatically label raw images. The automatically labeled data is then used together with the few ground-truthed images to train a regression neural network robust to label noise. Another work by Albert~\etal~\cite{2022_EGF_Irishunsuppre} instead uses an unsupervised learning algorithm~\cite{2021_ICLR_iMix} on the unlabeled data to learn better initial representations that allow for better accuracy numbers with limited amounts of labels. Finally, Skovsen~\etal~\cite{2021_SENSORS_skovsen2} published an updated version of their segmentation algorithm from synthetic data using style transfer GANs to simulate different weather conditions and multi-resolution prediction.

\subsection{Semi-supervised regression from images}
Semi-supervised regression (SSR) solves a regression task on a dataset where the labeled data is limited but the unlabeled data is plentiful. Although semi-supervised classification received many important contributions in the last years, the attention given to SSR has been limited.
Timilisina~\etal~\cite{2021_ASC_semigraphs} construct a fully connected graph from the feature representations of every sample before performing a bounded heat diffusion process to annotate the unlabeled data.
Jean~\etal~\cite{2018_NeurIPS_regkernels} adopt a Bayesian approach by fitting the labeled representations with gaussian processes and training an auxiliary regularization objective to minimize the predictive variance with regards to the unlabeled points.
Bzdok~\etal~\cite{2015_NeurIPS_semifactored} apply an autoencoder on top of medical images of brain voxels to solve a action regression task. The autoencoder is used to compress the input vectors and to ensure that the features extracted from labeled and unlabeled images will be compatible with the end logistic regressor.
Li~\etal~\cite{2017_AAAI_safepredreg} propose a process to aggregate the predictions from multiple regression predictions into a safe pseudo label for the unlabeled samples by means of solving a convex linear combination of each regressor output.
Zhou~\etal~\cite{2005_IJCAI_semicotrain} co-train two KNN regressors with different distance metrics that predict pseudo labels to be used by the other regressor on the unlabeled data, effectively reducing confirmation bias. Note that semi-supervised classification algorithms such as consistency regularization approaches~\cite{2019_NeurIPS_mixmatch,2019_IJCAI_ICT} or pseudo-labeling~\cite{2020_IJCNN_Pseudo} should translate to the regression setting.

\section{Unsupervised domain adaptation and super resolution on drone images}
We aim to solve the biomass prediction task jointly from a small set of ground-level images $\mathcal{X}_l$ with biomass labels $\mathcal{Y}_l$ (ground-level images) together with a large set of unlabeled (raw) images $\mathcal{X}_u$ from a different visual domain (drone images) in an unsupervised fashion. To do so, we use two neural networks: $\Psi$ performing super resolution and visual shift from the domain of $\mathcal{X}_u$ to $\mathcal{X}_l$ and $\Phi$, a regression network we use to learn jointly from $\mathcal{X}_l$ and $\mathcal{X}_u$ by optimizing a semi-supervised objective.
\begin{figure*}[t]
\centering{}\includegraphics[width=.6\linewidth]{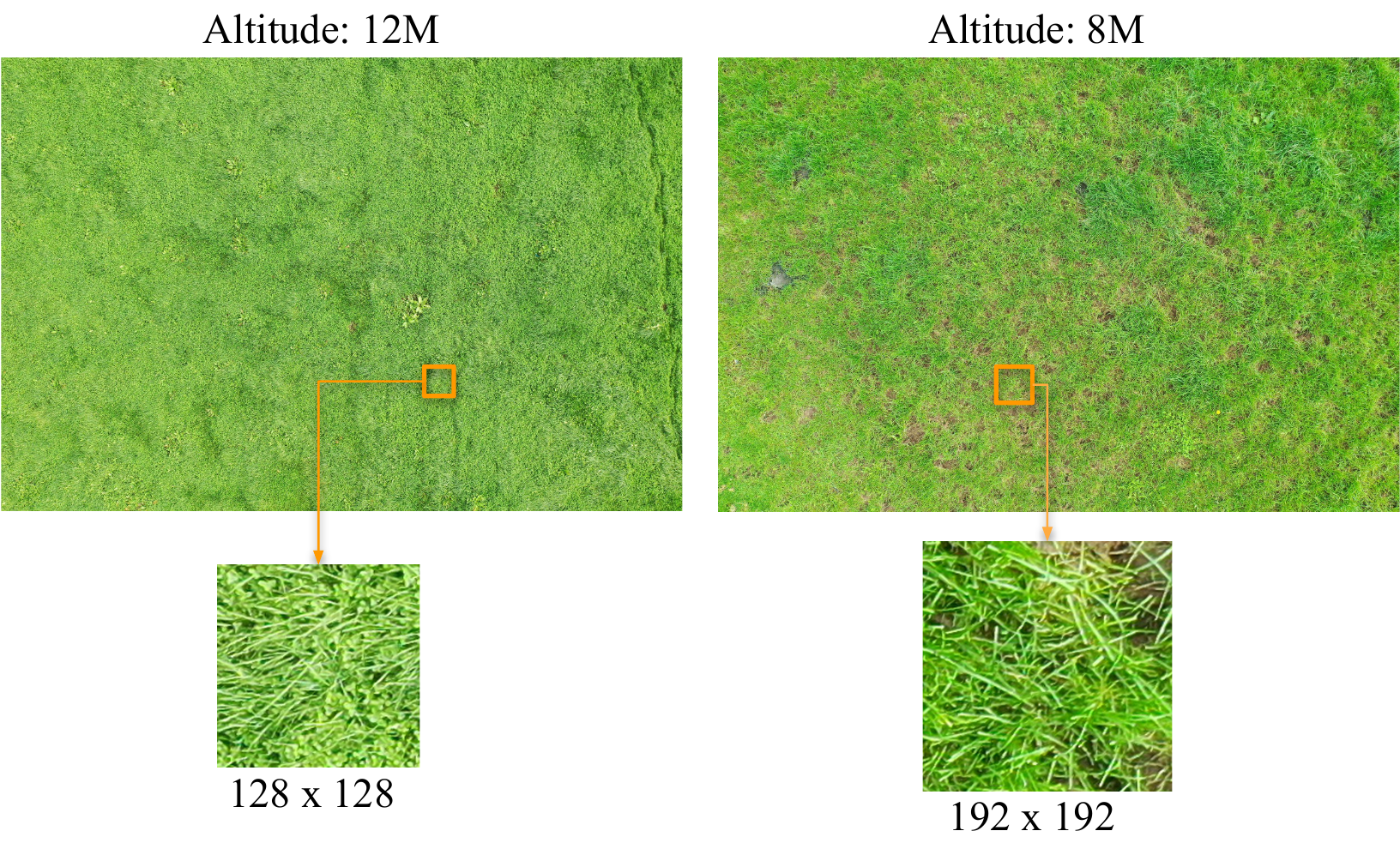}
\caption{Drone image cropping process at different altitudes. Given that resolution of the image is fixed, we increase or reduce the cropped area. All crops are then bicubicly upscaled to $2048\times 2048$ before deblurring.~\label{fig:diffaltitude}}
\end{figure*}

\subsection{Dataset presentation}
We consider here three different herbage biomass estimation datasets. The first one is the publicly available GrassClover dataset~\cite{2018_ICPA_grasscloverfirst}. This dataset is composed of $157$ annotated images (to be divided between training set and validation set) and $31.600$ unlabeled images. The image acquisition was carried out in Danish fields between 2017 and 2018 using for the most part an ATV mounted camera. The ground-truth collected is composed of the dry biomass percentages for the grass, white clover, red clover, total clover and weeds. The second dataset is the Irish clover dataset~\cite{2021_EGF_Irishdataset}, which is composed of $424$ training images, $104$ held out test images, and $594$ unlabeled images. The images were captured in the south of Ireland in the Summer of 2020 using a camera mounted on a tripod. The ground-truth collected is composed of the dry biomass percentages for grass, total clover and weeds (\%), the herbage height (cm), and the herbage dry matter per ha (kg DM/ha). Finally, we propose in this paper an extension of the Irish dataset where we collect drone images in the same 23 herbage paddocks originally studied in Ireland in late Autumn of 2021. We collect between $36$ and $7$ drone images per paddock at an altitude between $6$ and $12$ meters. The drone we use is the DJI Mavic 2 Pro~\footnote{https://www.dji.com/ie/mavic-2} with its default camera, taking pictures at a resolution of $5472\times 3648$. Although our drone is not capable of capturing its altitude relative to the land below, we subtract the above sea level GPS altitude of the drone from the land altitude at the associated GPS coordinates to obtain an approximate relative altitude using an open source API~\footnote{opentopodata.org}. We obtain $328$ drone images in total with their associated altitude. Because of the huge areas covered by drone images, the ground-truth we collect is limited to the dry herbage mass at the paddock level and we omit the grass height and biomass percentage information. The resulting $80$ labeled drone images are only used as a means to test the knowledge transfer from the ground-level to the drone images and not used for training. We propose two ground-truth estimations for the drone images: the first is a visual estimation performed on site at the time of the image collection by two human experts, very familiar with the site and that visually estimate the herbage on site every week. The second is obtained following the protocol of Egan~\etal~\cite{2018_JDS_incorporatingVM}: we cut two $1.2\times 8$ meters strips in the paddocks 4 cm above ground level (typical cow grazing height) using an Etesia lawn mower (Etesia UK. Ltd., Warwick, UK). A $100$ grams sample is collected from the cut material and dried at $95$°C for $16$ hours to obtain the dry herbage mass. We compare our algorithm against the human estimation and the exact ground-truth.
\begin{figure*}[t]
\centering{}\includegraphics[width=.75\linewidth]{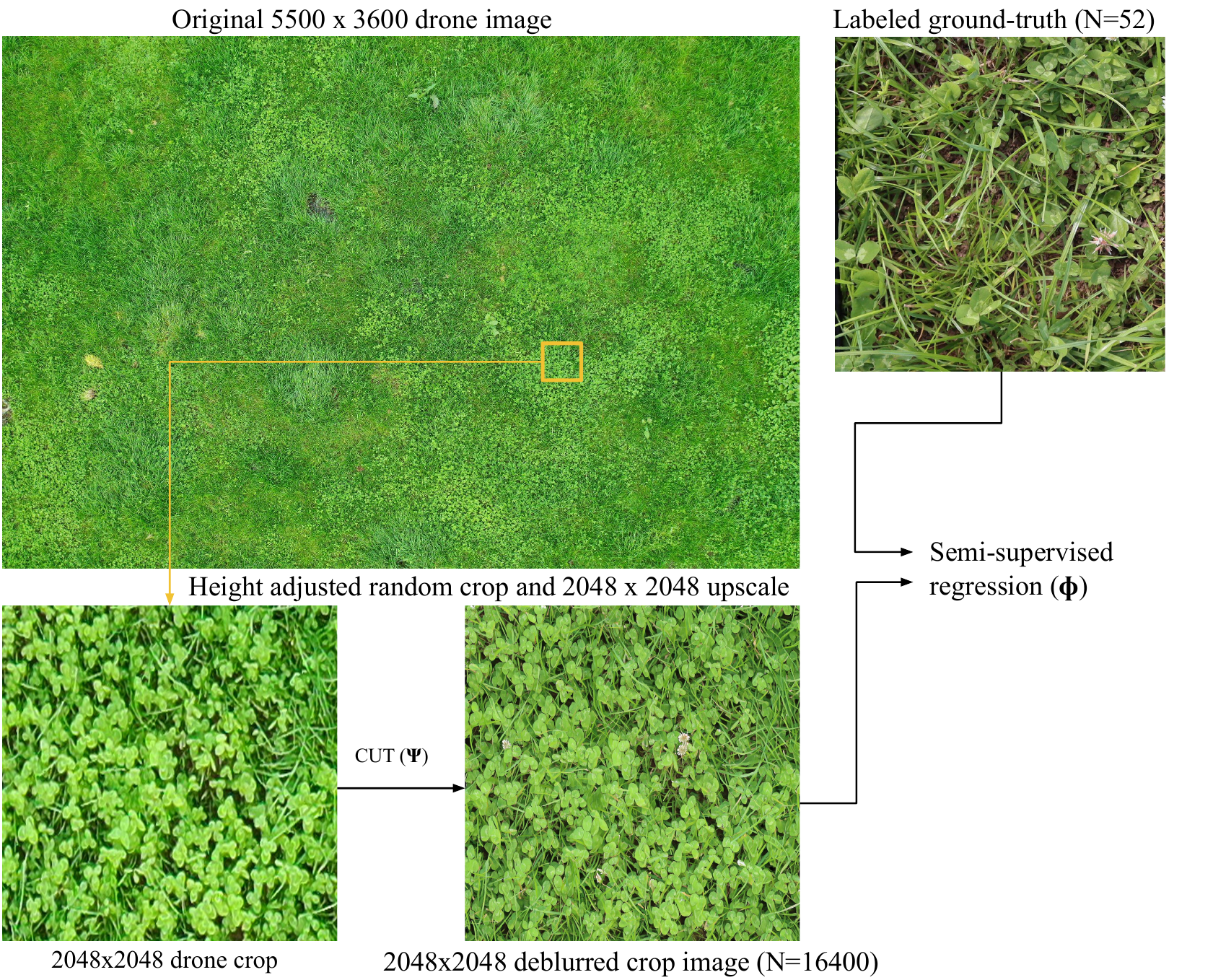}
\caption{Overview of our up-sampling and knowledge transfer algorithm. We use an up-sampling and visual domain transfer network $\Psi$ and a semi-supervised network $\Phi$ that we use to learn jointly from few labeled ground-level examples ($N=54$) and unlabeled drone images.~\label{fig:fullalgo}}
\end{figure*}
\subsection{Contrastive Unpaired Translation (CUT)}
The first step of our algorithm is to increase the resolution of drone images and to modify them to appear visually closer to the few ground-truthed images captured using high resolution cameras on the ground. To do so, we use Contrastive Unpaired Translation (CUT)~\cite{2020_ECCV_CUT}. CUT trains an adversarial network (GAN) to perform unpaired image style transfer using three principal components. $G$ is the generator part of the network, competing to fool $D$ the discriminator in an alternative adversarial optimization and $F$ the projection head is used to optimize the contrastive part of the algorithm, which promotes semantic similarities between the same image before and after the visual transformation. CUT minimizes a combination of three losses to learn the parameters for $G$, $D$, and $F$. First the adversarial loss~\cite{2014_NeurIPS_GAN}
\begin{equation}
\begin{aligned}
    \mathbf{L}_{adv}(G,D,\mathcal{X}_l,\mathcal{X}_u) & = \mathbb{E}_{x_l\sim\mathcal{X}_l}\log D(x_l) \\
    & + \mathbb{E}_{x_u\sim\mathcal{X}_u}(1 - \log D(G(x_u))),
\end{aligned}
\end{equation}
promotes the generator $G$ to transform images from $\mathcal{X}_u$ (the drone images) so that they become indistinguishable by the discriminator $D$ from the high resolution ground-level images in $\mathcal{X}_l$.
Second, once the image has been transformed by the generator, a patch contrastive regularization objective is applied where patches at the same location in the image before and after the transformation are encouraged to have similar features after projection through $F$ while being dissimilar to any other random patch from the image. This results in a constrastive patch objective
\begin{equation}
    \mathbf{L}_{patch}(G,F,X) = - \frac{1}{P}\sum_{i=1}^P\log\left(\frac{\exp{\left(\textit{ip}(p_i, p_i')/\tau\right)}}{\sum_{k=1}^P\exp{\left(\textit{ip}(p_k, p_i')/\tau\right)}}\right),
\end{equation}
where $P=64$ random patches are cropped out from the input image, their feature representations encoded through $G$ (stopping half way), projected through $F$ and $L2$ normalized. The process is repeated for the transformed version of the image to form $P$ pairs of random patches $\{(p_i, p_i')\}_{i=1}^P$ for a given image $x\in \mathcal{X}_u$ where $p_i$ is the representation before the domain shift and $p_i'$ after. The dot product between the representations of corresponding pairs is encouraged to be close to one and close to zero for different patches. $\mathbf{L}_{patch}$ can also be applied to images in $\mathcal{X}_l$ to enforce that $G$ will perform the identity operation, \ie $\forall x\in\mathcal{X}_l, G(x) = x$. $\tau = 0.07$ is the contrastive temperature parameter. The final objective minimized by CUT where $\lambda_1 = \lambda_2 = 0.5$ is
\begin{equation}
\begin{aligned}
    \mathbf{L} & = \mathbf{L}_{adv}(G,D,\mathcal{X}_l,\mathcal{X}_u) \\  
    & + \lambda_1\mathbf{L}_{patch}(G,F,\mathcal{X}_u) +\lambda_2\mathbf{L}_{patch}(G,F,\mathcal{X}_l).
\end{aligned}
\end{equation}

\subsection{CUT for super resolution and style transfer}
\paragraph{Cropping the drone images.} We propose to crop squared areas from the drone images to obtain similar amounts of elements per image as ground level images. Because the drone images were not all captured at the same altitude, we adjust the area cropped out from the drone images depending on the altitude at which the image was captured. We observe visually that at an altitude of 8 meters, a $256\times 256$ pixel crop of the drone data yields similar numbers of grass elements and of similar size to the ground-level images. Given the altitude of the drone at the time the picture was taken, we multiply the edge of the crop by the ratio between the altitude and the standard value of 6 meters~\ie for an altitude of 12 meters, the edge of the square crop will be $6/12\times256 = 128$. Figure~\ref{fig:diffaltitude} illustrates the height adjusted cropping process. This process allows us to capture the same area of land independently of the height of the drone.
\paragraph{Deblurring the crops}
Although CUT is originally designed to transfer styles between two unpaired visual domains, we propose here to task the algorithm with improving the resolution of the drone images while at the same time transferring their visual style to ground-level images. Note that the super-resolution task is usually performed by conditional GANs (e.g.~\cite{2020_ECCV_HAN}) but we propose here to use an unpaired algorithm. For each image $x\in\mathcal{X}_u$, we upscale the image from the original resolution to $2048\times2048$. $\Psi$ is then trained to transfer the visual style of the ground-level high resolution images to the up-sampled crops, effectively deblurring them to appear closer to the higher resolution images (see Figure~\ref{fig:upsamplingexample}).

\subsection{Semi-supervised regression on drone data}
By up-sampling and visually transforming drone images to appear closer to the ground-level visual domain, we are now able to learn jointly from $\mathcal{X}_l$ and $\mathcal{X}_u$. Since it is only practical to obtain labels for ground-level camera images, we propose to optimize a semi-supervised regression objective using $\mathcal{X}_l$ as the labeled set and $\mathcal{X}_u$ as the unlabeled data. After an initial pretraining of $\Phi$ on $\mathcal{X}_l$, we start guessing biomass labels for ${X}_u$ using a consistency regularization approach~\cite{2019_NeurIPS_mixmatch}. Using two data augmented views $x_u'$ and $x_u''$ (vertical and horizontal random flipping), we use an exponential moving average (EMA) on the weights of $\Phi$ to guess two approximate biomass labels $y_u'$ and $y_u''$ for $x_u$. Rather than averaging the two approximate labels with equal importance like consistency regularization algorithms for image classification~\cite{2019_NeurIPS_mixmatch,2020_ICLR_ReMixMatch}, we draw a random mixing parameter $\lambda$ from a uniform distribution to improve the regularization of the predictions and avoid confirmation bias~\cite{2020_IJCNN_Pseudo}. We obtain an approximate label $\Tilde{y} = \lambda y_u' + (1-\lambda) y_u''$ for every unlabeled images. 
We enforce the distribution of the predictions on unlabeled samples to match the observed ground-truth distribution on the labeled data (distribution alignment) by multiplying the label prediction by the ratio between a sliding window average ($50$ mini-batches in practice) of $\Tilde{y}$ and the observed distribution on the labeled data. EMAs and distribution alignment are common principles of consistency regularization algorithms for semi-supervised learning~\cite{2020_ICML_MentorMix,2020_ICLR_ReMixMatch}. Finally, we normalize the biomass composition to sum to 1 in the approximate label $\Tilde{y}$. Figure~\ref{fig:fullalgo} presents an overview of the proposed semi-supervised training algorithm

\subsection{Regression from images}
We predict biomass labels from images in $\mathcal{X}_l$ and $\mathcal{X}_u$ using $\Phi$ to extract visual features. For the Irish dataset~\cite{2021_EGF_Irishdataset}, we use three different linear heads, separating the predictions of the herbage mass, herbage height, and biomass composition. We normalize the herbage mass and herbage height values between 0 and 1 using fixed normalization values ($4000$ kg DM/ha for the herbage mass and $20$cm for the height) and offset them by $+0.2$ to improve the prediction for low values originally too close to $0$. To obtain values between zero and one for each target prediction and ensure that the sum of the biomass percentages equals one, we apply a softmax function on the three outputs from the biomass head and a sigmoid function for each of the other two values. For the GrassClover dataset~\cite{2019_CVPRW_grasscloverdataset}, we use a single linear head, predicting the biomass percentages (grass, white clover, red clover, weeds) and sum the predictions for white and red clover to obtain the total clover content. These configurations follow the work of Albert~\etal~\cite{2021_ICCVW_semisupmine}.
We use the root mean squared error (RMSE) as the training objective.

\begin{figure}[t]
\centering\includegraphics[width=.8\linewidth]{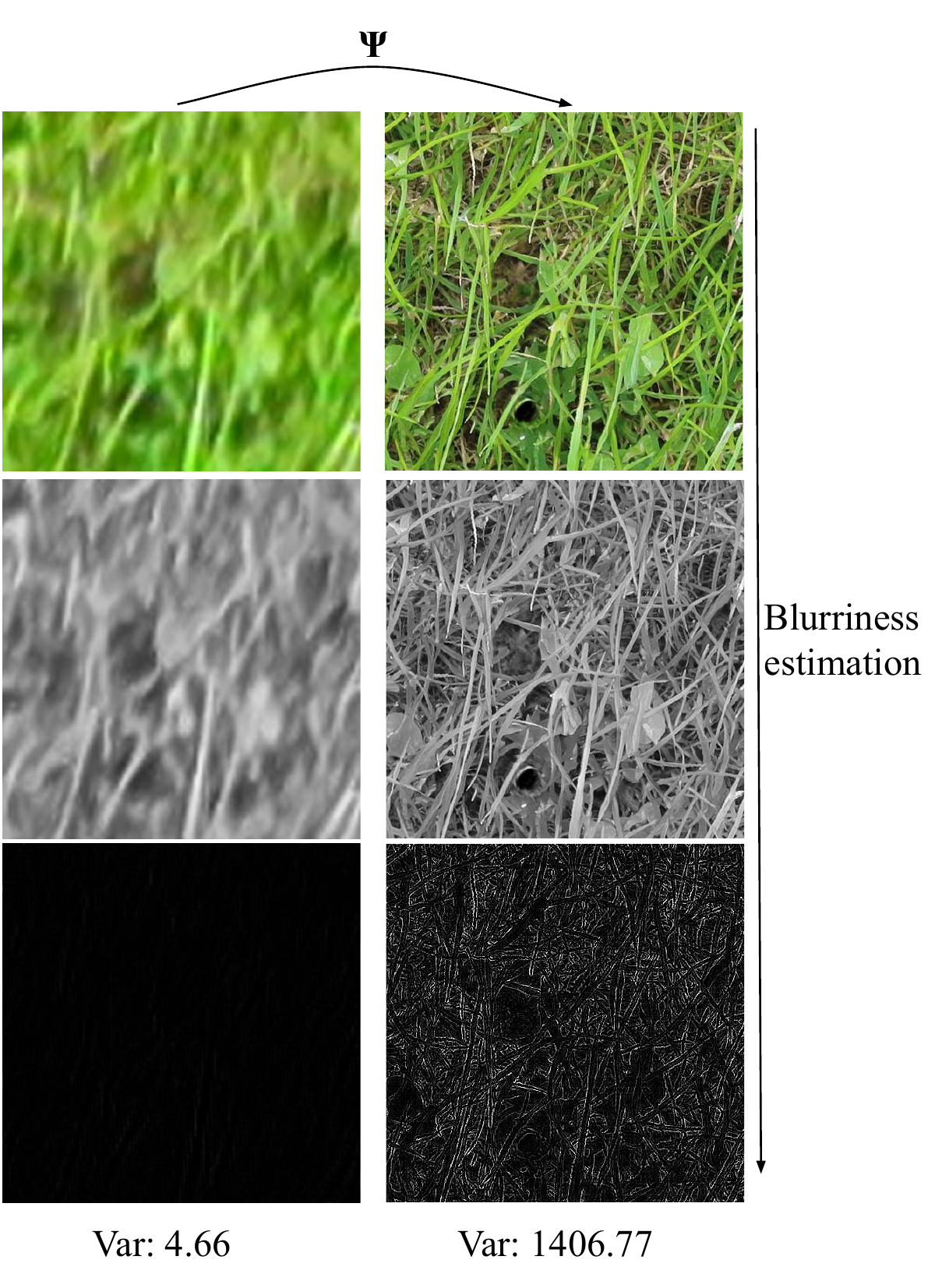}
\caption{Overview of the deblurring effect of $\Psi$ on drone data. A high variance indicates a sharper image.\label{fig:blur}}
\end{figure}

\begin{table*}[]
    \global\long\def\arraystretch{1}%
    \centering
    \resizebox{.95\textwidth}{!}{{{}}%
    \begin{tabular}{l>{\centering}c>{\centering}c>{\centering}c>{\centering}c>{\centering}c>{\centering}c>{\centering}c>{\centering}c>{\centering}c>{\centering}c>{\centering}c}
    \toprule
        & \multicolumn{5}{c}{HRMSE} & & \multicolumn{4}{c}{RMSE} \tabularnewline
        \cmidrule(lr){2-6}\cmidrule(lr){8-11}   
        & Total & Grass & Clover & Weeds & Avg. & HRE & Grass & Clover & Weeds & Avg. & HE \tabularnewline
        \midrule
        Albert~\etal~\cite{2021_ICCVW_semisupmine} & 230.10 & 220.84 & 34.86 & 27.13 & 94.28 & 1.14 & 4.81 & 4.75 & 3.42 & 4.33 & 2.15 \tabularnewline
        Albert~\etal~\cite{2022_EGF_Irishunsuppre} & 229.12 & 218.02 & 37.65 & 29.21 & 94.96 & 1.09 & \textbf{4.58} & 4.22 & 3.44 & \textbf{4.08} & \textbf{2.03} \tabularnewline
        \midrule
        Labeled only & 229.23 & 268.90 & 107.39 & 39.82 & 138.71 & 1.08 & 17.15 & 14.08 & 4.74 & 11.99 & 2.28 \tabularnewline
        Semi-sup & 234.50 & 224.10 & 43.03 & 26.74 & 97.96 & 1.04 & 5.85 & 5.51 & \textbf{3.19} & 4.85 & 2.24 \tabularnewline
        + distribution alignment & 220.79 & 215.88 & 40.00 & 26.78 & 94.22 & 1.08 & 5.53 & 5.51 & 3.25 & 4.76 & 2.09 \tabularnewline
        + EMA & 217.28 & 208.96 & 34.16 & \textbf{26.50} & 89.88 & 1.08 & 4.86 & 4.73 & 3.26 & 4.28 & 2.09 \tabularnewline
        + Self-sup~\cite{2021_ICLR_iMix} & 211.57 & 202.18 & \textbf{28.93} & 26.80 & \textbf{85.97} & 1.09 & 4.54 & 4.50 & 3.21 & \textbf{4.08} & 2.09 \tabularnewline
        drone unlabeled & \textbf{209.69} & \textbf{199.61} & 33.59 & 27.13 & 86.78 & \textbf{1.02} & 4.74 & 4.65 & 3.33 & 4.24 & 2.18 \tabularnewline
       \bottomrule
    \end{tabular}}
    \caption{Ablation study and comparison against state-of-the-art algorithms on the Irish dataset. The last row denotes replacing the ground-level unlabeled camera images with deblurred drone images. The best results are in bold.\label{tab:irishdataset}}
\end{table*}

\section{Experiments}
\subsection{Experimental setup}
We conduct experiments on two biomass prediction datasets from canopy images. For the GrassClover dataset, we use $100$ labeled and $1,000$ unlabeled images for training and $57$ images for validation. We report the test accuracy results on the evaluation server for the GrassClover dataset. For the Irish dataset, we use $52$ labeled and $595$ unlabeled images for training and $104$ images for validation. For the drone images, we extract 50 random crops from each of the $328$ images to create a dataset of $16,400$ unlabeled images. We train using stochastic gradient descend at a resolution of $512\times512$ with a batch size of $32$ and a fixed learning rate of $0.03$. We update the EMA with a multiplication parameter of $0.99$ at every mini-batch. The training augmentations are resize, random crop, random horizontal and vertical flipping, and normalization. When we perform semi-supervised learning, we create each mini-batch by aggregating $4$ labeled samples with unlabeled images as in Albert~\etal~\cite{2021_ICCVW_semisupmine}. For the neural networks, we use a ResNet18~\cite{2016_CVPR_ResNet} pretrained on ImageNet~\cite{2012_NeurIPS_ImageNet} for the regression network $\Phi$ and the 9 ResNet blocks version of the CUT model $\Psi$.

\subsection{Drone image deblurring and style transfer}

We evaluate the deblurring capacity of $\Psi$ by computing the variance of the Laplacian on the grayscale view of an image. Computing the Laplacian of the image allows us to extract edges in the image and the variance of the resulting value quantifies the sharpness of the edges: $\mathrm{sharpness} = \mathbf{var}(\mathrm{\nabla^2}(\mathrm{grayscale}(\mathrm{image})))$~\cite{2000_ICPR_blurr} with $\mathrm{\nabla^2}$ the Laplacian operator. A higher variance indicates a better defined (sharper) image. The sharpness estimation process is illustrated in Figure~\ref{fig:blur}. We observe that the variance averaged over all the crops changes from $5.32$ when cropping directly from the drone images to $1261.05$ for the same images deblurred by $\Psi$.
\begin{table}[]
    \global\long\def\arraystretch{1}%
    \centering
    \resizebox{.3\textwidth}{!}{{{}}%
    \begin{tabular}{l>{\centering}c>{\centering}c>{\centering}c>{\centering}c>{\centering}c>{\centering}c>{\centering}c>{\centering}c}
    \toprule
        & HRMSE & HRE \tabularnewline
        \cmidrule(lr){1-3}
        \multicolumn{3}{c}{Against harvested ground-truth} \tabularnewline
        \cmidrule(lr){1-3}
        Labeled only & 1094.18 & 0.43\tabularnewline
        Semi-sup. & 566.68 & 0.81 \tabularnewline
        Semi-sup. drone & 219.15 & 0.97 \tabularnewline
        \midrule
        Human expert & 170.03 & 1.04 \tabularnewline
       \bottomrule
    \end{tabular}}
    \caption{Results on drone images. Errors are computed against the absolute harvested ground-truth.\label{tab:resultsdrone}}
\end{table}

\begin{table*}[]
    \global\long\def\arraystretch{1}%
    \centering
    \resizebox{.65\textwidth}{!}{{{}}%
    \begin{tabular}{@{}>{\centering}c>{\centering}c>{\centering}c>{\centering}c>{\centering}c>{\centering}c>{\centering}c}
    \toprule
        \cmidrule(lr){1-6}
        \multicolumn{6}{c}{HRMSE} \tabularnewline
        \cmidrule(lr){1-6}
        1 & 5 & 20 & 50 & 100 & all \tabularnewline
        \midrule
        $252.71\pm51.24$ & $227.18\pm20.15$ & $222.97\pm12.56$ & $219.15\pm6.93$ & $220.09\pm4.14$& $219.53$  \tabularnewline
       \bottomrule
    \end{tabular}}
    \caption{RMSE errors on drone image for varying amounts of random crops per image for the best performing model. Averaged over 5 random sets of crops.\label{tab:resultsncrops}}
\end{table*}
\begin{figure}[t]
\centering{}\includegraphics[width=.9\linewidth]{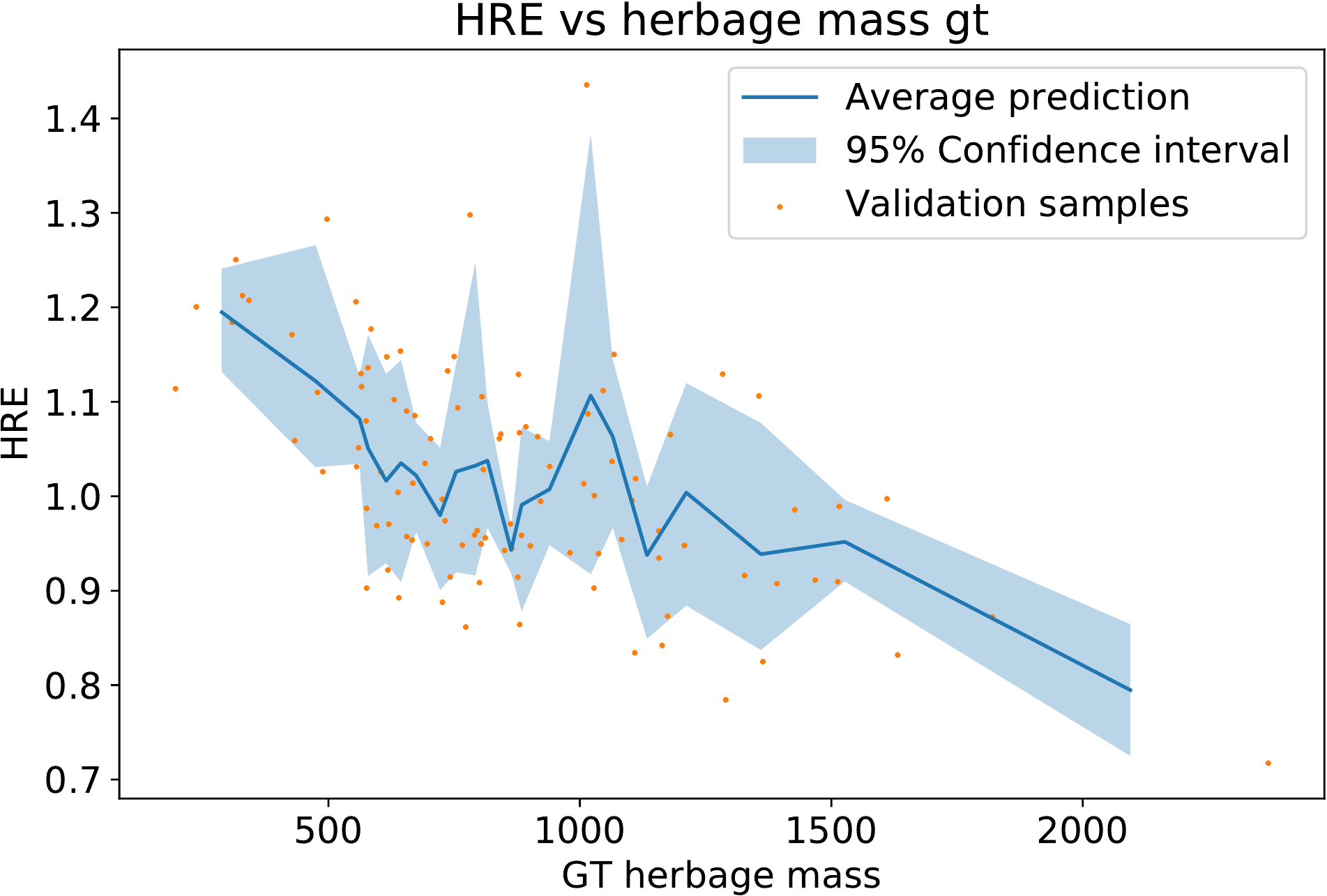}
\caption{Visualization of the HRE on the validation set of the Irish dataset with 95\% confidence intervals.~\label{fig:hreline}}
\end{figure}
\begin{table}[]
    \global\long\def\arraystretch{1}%
    \centering
    \resizebox{.5\textwidth}{!}{{{}}%
    \begin{tabular}{l>{\centering}c>{\centering}c>{\centering}c>{\centering}c>{\centering}c>{\centering}c>{\centering}c>{\centering}c}
    \toprule
        &  &  \multicolumn{3}{c}{Clover} & &  \tabularnewline
        \cmidrule(lr){3-5}
        & Grass & Total & White & Red & Weeds & Avg. \tabularnewline
        \midrule
        Skovsen~\etal~\cite{2019_CVPRW_grasscloverdataset} & 9.05 & 9.91 & 9.51 & 6.68 & 6.50 & 8.33 \tabularnewline
        Naranayan~\etal~\cite{2020_IMVIP_extracting} & 8.64 & 8.73 & 8.16 & 10.11 & 6.95 & 8.52 \tabularnewline
        Albert~\etal~\cite{2021_ICCVW_semisupmine} & 8.78 & 8.35 & \textbf{7.72} & \textbf{7.35} & 7.17 & 7.87 \tabularnewline
        \midrule
        Labeled only & 9.81 & 8.49 & 7.99 & 8.58 & 7.25 & 8.57 \tabularnewline
        Semi-sup. & \textbf{6.68} & \textbf{7.76} & 8.08 & 8.66 & \textbf{6.72} & \textbf{7.58} \tabularnewline
       \bottomrule
    \end{tabular}}
    \caption{Results on the GrassClover test set (RMSE). Lowest errors are in bold.\label{tab:resultsdanish}}
\end{table}
\subsection{Semi-supervised biomass, herbage height and herbage mass prediction}
We evaluate the capacity of our algorithm to predict the biomass composition of herbage (\%) together with an estimation of the dry herbage mass (kg DM/ha) and the grass height (cm) in a semi-supervised manner. We run experiments on the Irish dataset with the original unlabeled images but also study replacing them with an equal number ($N=596$) of deblurred drone images. This is to evaluate the capacity for drones to capture unlabeled data that can be used to improve the prediction at the ground-level. For evaluation purposes, we compute the Herbage Root Mean Square Error (HRMSE) which is the RMSE between predicted and ground-truth herbage mass for grass, clover, weeds and for the total mass and the Herbage Relative Error (HRE) which is the ratio between the ground-truth value of the total herbage mass and the prediction: $\mathit{HRE} = \frac{\mathit{pred}_{\mathit{herbage}}}{\mathit{gt}_\mathit{herbage}}$. The $\mathit{HRE}$ measure is typically used to compare human visual estimation against the collected ground-truth~\cite{2018_JDS_incorporatingVM}. We additionally compute the RMSE over the predicted percentages of grass, clover, weeds in the herbage and the Height Error (HE) which is the RMSE between the predicted herbage height and ground-truth height. We compare against state-of-the-art results on the Irish dataset where the validation images are ground-level images in Table~\ref{tab:irishdataset}. We study the importance of the different elements of the semi-supervised algorithm on the validation error, including enforcing distribution alignment for the label guesses and using an exponential moving average (EMA) on the weights of the semi-supervised network. We also report results when initializing the weights of the network using an unsupervised representation learning algorithm~\cite{2021_ICLR_iMix} on the unlabeled data as in Albert~\etal~\cite{2022_EGF_Irishunsuppre}. We finally point out that using an equal number of deblurred drone images produces comparable results to using the original unlabeled images (last two rows in Table~\ref{tab:irishdataset}). This result motivates the use of drone images to easily capture large amounts of unlabeled images. Figure~\ref{fig:hreline} shows a line plot of the HRE compared against the ground-truth herbage mass where we observe that the algorithm struggles the most on high or low herbage mass outliers ($<500$ to $>2000$ kg DM/ha). This is most likely due to the low amount of high or low herbage mass examples seen during training.

\subsection{Prediction on drone images}
Table~\ref{tab:resultsdrone} reports the Herbage Root Mean Square Error (HRMSE) when predicting on drone data without the need to gather additional labels. First, we evaluate the accuracy of a CNN model learned on ground-level data only to predict on deblurred drone patches using $\Psi$. We then evaluate the accuracy benefits of training on deblurred drone patches in a semi-supervised manner for the herbage mass prediction and compare the error rate of our algorithm against human experts. We observe a significant reduction in error rates when the drone images are used in the semi-supervised objective (semi-sup.~drone row) over using only the ground-level data (labeled only or semi-sup.~with the unlabeled ground-level images). The performance proposed by our low supervision algorithm is close to be on par with human experts at the paddock level.

Table~\ref{tab:resultsncrops} reports on how augmenting the number of random crops improves the prediction of the best performing model. We report the average herbage mass error and standard deviation over 5 random sets of crops. ``All" denotes cropping the image in a checkerboard fashion and using all crops (from $250$ to $1,000$ crops per image depending on the altitude). Although 1 random crop per image yield interestingly good results, we validate our choice of 50 crops per images for more stable predictions.

\subsection{Semi-supervised biomass prediction on the GrassClover dataset}
We compare our semi-supervised approach against state-of-the-art algorithms on the publicly available GrassClover dataset in Table~\ref{tab:resultsdanish} where we report RMSE errors for the biomass percentage prediction on the held out test set\footnote{https://competitions.codalab.org/competitions/21122} where we perform on par with existing approaches.

\section{Conclusion}
This paper investigates how to extend the biomass estimation and herbage mass prediction problem from ground-level to drone images. By its nature, the herbage biomass information of drone images is hard to annotate finely because of the huge areas covered. Ground-level data, however, has the advantage of providing easier to acquire, finely annotated, and high resolution images of the herbage but would not be a good solution to generalize to targeted fertilization on entire herbage fields. To successfully transfer knowledge from the ground-truth images to the drone data, we propose to train an unpaired style transfer algorithm to deblur height adjusted crops of drone images, increase resolution by a factor of 8 and to transfer the visual style of ground-level images, captured using different cameras, to look more similar to their ground-level counterparts. The large set of transformed unlabeled drone images is used together with the finely annotated ground-level images to learn unsupervised initialization weights and to train a semi-supervised regression algorithm. The neural network trained on the partially labeled set largely improves the regression accuracy on the ground-level data and the herbage mass prediction on drone images. We significantly reduce the prediction gap compared to a human expert, achieving error rates close to experienced technicians familiar with the land with little extra annotation cost than few ground-level images. This challenge prompts further research in the field of low supervision computer vision for herbage biomass prediction from drone images. Future work would involve using the approximate predictions made by the human experts to improve the results and a smart algorithm capable of sampling areas of interest in the drone images to be predicting upon by the herbage biomass algorithm. 
\section*{Acknowledgements}
This publication has emanated from research conducted with the financial support of Science Foundation Ireland (SFI) under grant number SFI/15/SIRG/3283, SFI/16/RC/3835 and SFI/12/RC/2289\_P2 and the computing support of the Irish Centre for High End Computing (ICHEC).

{\small
\bibliographystyle{ieee_fullname}
\bibliography{egbib}
}

\end{document}


\title{Supplementary marterial for Unsupervised domain adaptation and super resolution on drone images for autonomous dry herbage biomass estimation}
\author{Paul Albert$^{1,3,5}$, Mohamed Saadeldin$^{2,3,5}$, Badri Narayanan$^{2,3,5}$, Jaime Fernandez$^{1,3}$, Brian Mac Namee$^{2,3,5}$, \\ Deirdre Hennessy$^{4,5}$, Noel E. O{'}Connor$^{1,3,5}$, Kevin McGuinness$^{1,3,5}$\\\\
$^1$School of Electronic Engineering, Dublin City University\\ $^2$School of Computer Science, University College Dublin\\ $^3$Insight Centre for Data Analytics \\ $^4$Teagasc, $^5$VistaMilk\\
{\tt\small paul.albert@insight-centre.org}
}
\maketitle
{\small
\bibliographystyle{ieee_fullname}
\bibliography{egbib}
}
\section{Prediction histograms on drone images}
\begin{figure}[b]
\centering{}\includegraphics[width=\linewidth]{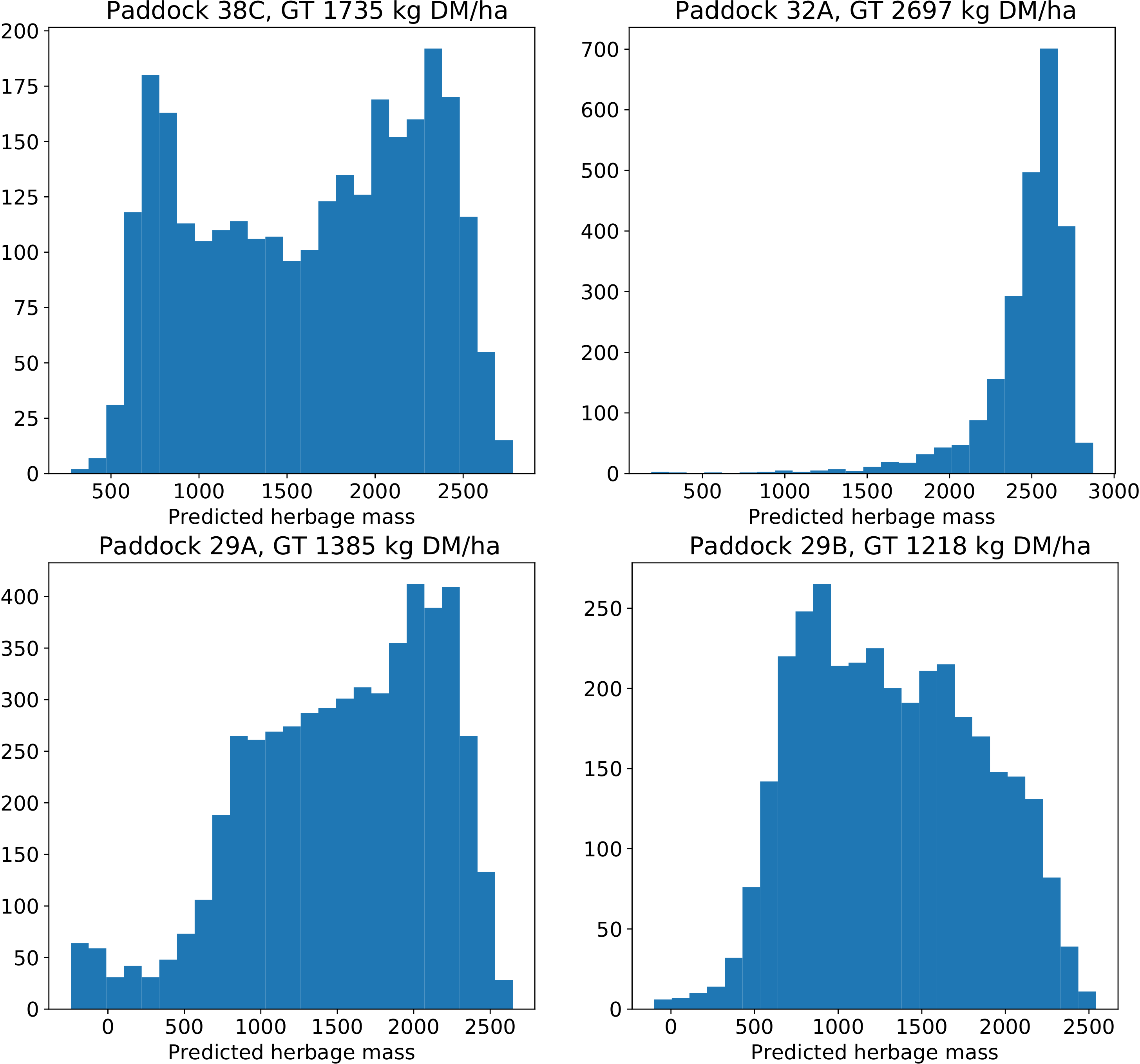}
\caption{Histograms of the prediction on crops of drone images from a paddock.~\label{fig:histo}}
\end{figure}

Figure~\ref{fig:histo} displays histograms of the herbage mass prediction for all the crops from the drone images of a given paddock. We find that although some outliers exist, the mode of the predictions is often close the actual ground-truth. These histograms open the research question of creating an algorithm capable of selecting crops from relevant areas of the drone images which would be more representative of the actual ground-truth.

\section{Herbage prediction from drone image}
\begin{figure*}[t]
\centering{}\includegraphics[width=\linewidth]{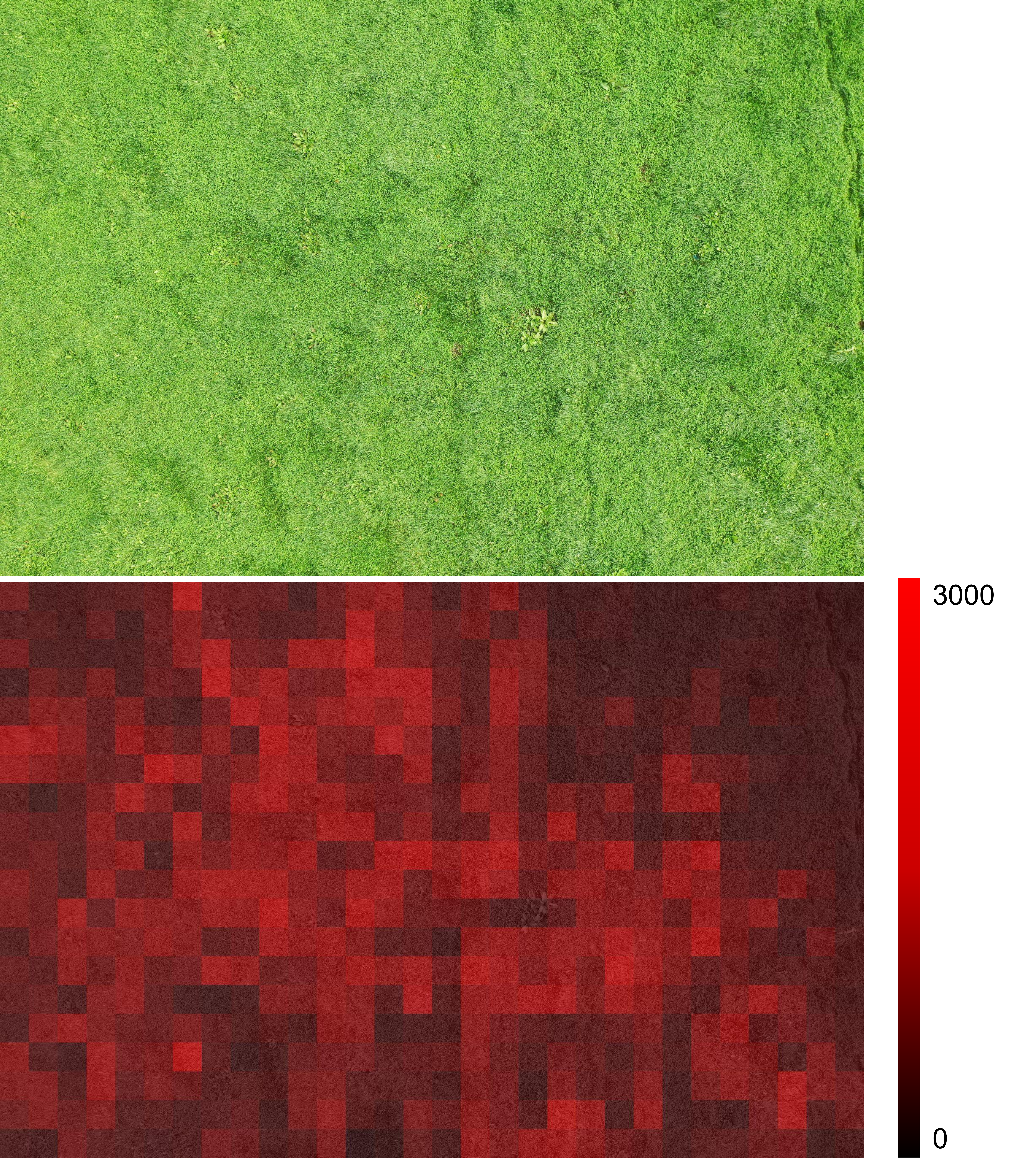}
\caption{Prediction on ordered crops of the drone images. Paddock level herbage mass ground-truth: $1624$ kg DM/ha, average prediction: $1347.45$ kg DM/ha. Altitude $9.3$ meters.~\label{fig:drone1}}
\end{figure*}

Figure~\ref{fig:drone} and Figure~\ref{fig:drone1} display the herbage mass prediction results for a drone image for each ordered crops. The ground level resolution is approximately $0.5\times 0.5$ meters.

\begin{figure*}[t]
\centering{}\includegraphics[width=\linewidth]{latex/108.pdf}
\caption{Prediction on ordered crops of the drone images. Paddock level herbage mass ground-truth: $1735$ kg DM/ha, average prediction: $1719.21$ kg DM/ha. Altitude $8.2$ meters.~\label{fig:drone}}
\end{figure*}

